\documentclass[11pt]{article}
\usepackage{acl2016}
\usepackage{times}
\usepackage{soul,color}
\usepackage{url}
\usepackage{latexsym}
\usepackage{graphicx}
\usepackage{color}
\usepackage{xcolor}
\usepackage{tikz}
\usepackage{amsmath}
\usepackage{caption}
\usepackage{paralist}
\usepackage{subfig}
\usepackage{booktabs}
\usepackage{amssymb}
\usepackage{amsmath}
\usepackage{enumerate}
\usepackage{xspace}
\usepackage{multirow}
\usepackage{paralist,relsize}
\usepackage{microtype}

\aclfinalcopy 
\def\aclpaperid{655} 

\newcommand{\tabref}[2][]{Table~\ref{#2}\xspace}
\newcommand{\figref}[2][]{Figure~\ref{#2}\xspace}
\newcommand{\secref}[2][]{\S\ref{#2}\xspace}

\newcommand{\relation}[2][]{\textsc{#2}$_{\text{#1}}$\xspace}
\newcommand{\vect}[2][]{\ensuremath{\textit{#2}_{\mathit{#1}}\xspace}}

\newcommand{\diffvec}[1][]{\textsc{DiffVec}#1\xspace}

\newcommand{\nounsp}{\relation[SP]{Noun}}
\newcommand{\verbrel}{\relation{Verb}}
\newcommand{\verbthree}{\relation[3]{Verb}}
\newcommand{\verbthreepast}{\relation[3Past]{Verb}}
\newcommand{\verbpast}{\relation[Past]{Verb}}
\newcommand{\verbnoun}{\relation{VerbNoun}}
\newcommand{\lvc}{\relation{LVC}}
\newcommand{\prefix}{\relation{Prefix}}
\newcommand{\nouncoll}{\relation[Coll]{Noun}}

\newcommand{\attribute}{\relation[Attr]{LexSem}}
\newcommand{\causepurp}{\relation[Cause]{LexSem}}

\newcommand{\refrel}{\relation[Ref]{LexSem}}
\newcommand{\spacerel}{\relation[Space]{LexSem}}

\newcommand{\eventrel}{\relation[Event]{LexSem}}
\newcommand{\hyperrel}{\relation[Hyper]{LexSem}}
\newcommand{\merorel}{\relation[Mero]{LexSem}}

\newcommand{\oppos}{{\vect[w1,w2]{Oppos}}}
\newcommand{\shuff}{{\vect[w1,w2]{Shuff}}}

\newcommand{\postag}[1]{{\tt #1}\xspace}

\newcommand{\bst}{\bf}

\newcommand{\myurl}[1]{{\smaller\url{#1}}\xspace}

\newcommand{\classtype}[1]{\textsc{#1}\xspace}
\newcommand{\closedworld}{\classtype{Closed-World}}
\newcommand{\openworld}{\classtype{Open-World}}

\newcommand{\wvec}[1]{\ensuremath{\mathbf{#1}\xspace}}

\newcommand{\z}{\phantom{0}}

\newcommand{\method}[2][]{\texttt{#2}$_{\text{#1}}$\xspace}
\newcommand{\wordveclong}{\method{word2vec}}
\newcommand{\wordvec}[1][]{\method[#1]{w2v}}
\newcommand{\glove}[1][]{\method[#1]{GloVe}}

\newcommand{\hlbl}{\method{HLBL}}
\newcommand{\senna}{\method{SENNA}}

\newcommand{\svd}[1][]{\method[#1]{SVD}}

\newcommand{\wiki}{wiki\xspace}

\newcommand{\precision}[1][]{\ensuremath{\mathcal{P}_{#1}}\xspace}
\newcommand{\recall}[1][]{\ensuremath{\mathcal{R}_{#1}}\xspace}
\newcommand{\fscore}[1][]{\ensuremath{\mathcal{F}_{#1}}\xspace}

\newcommand{\withneg}{\ensuremath{+neg}}

\newcommand{\lex}[1]{\textit{#1}\xspace}

\newcommand{\wordpair}[2]{\ensuremath{(\lex{#1},\lex{#2})}\xspace}

\newcommand\ws{\textbf{w}}
\newcommand\wsub[1]{\ws_{#1}}
\newcommand\wi{\wsub{i}}

\newcommand\wk{\wsub{k}}
\newcommand\wij{\wsub{i-j}}
\newcommand\twsub[1]{\tilde{\ws}_{#1}}
\newcommand\twi{\twsub{i}}
\newcommand\twj{\twsub{j}}
\newcommand\twij{\twsub{i+j}}

\newcommand{\nlp}{\textsc{nlp}\xspace}

\newcommand{\spadeaff}{\ensuremath{1}\xspace}
\newcommand{\clubaff}{\ensuremath{2}\xspace}

\title{\lex{Take} and \lex{Took}, \lex{Gaggle} and \lex{Goose},
  \lex{Book} and \lex{Read}: Evaluating the
  Utility of Vector Differences for Lexical Relation Learning}

\author{Ekaterina Vylomova,$^{\spadeaff}$ Laura Rimell,$^{\clubaff}$ Trevor Cohn,$^{\spadeaff}$ \and Timothy Baldwin$^{\spadeaff}$\\ 
  $^{\spadeaff}$Department of Computing and Information Systems, University of Melbourne\\
   $^{\clubaff}$Computer Laboratory, University of Cambridge\\
  \texttt{\small evylomova@gmail.com laura.rimell@cl.cam.ac.uk \{tcohn,tbaldwin\}@unimelb.edu.au}\\[-0.0ex]}

\begin{document}

\maketitle

\renewenvironment{abstract}
 {
  \begin{center}
  \bfseries \abstractname\vspace{-.5em}\vspace{0pt}
  \end{center}
  \list{}{
    \setlength{\leftmargin}{0.4cm}%
    \setlength{\rightmargin}{\leftmargin}%
  }%
  \item\relax}
 {\endlist}

\begin{abstract}
  Recent work 
has shown that simple vector
  subtraction over word embeddings
  is surprisingly effective at capturing different lexical relations,
  despite lacking explicit supervision. Prior work has evaluated this
  intriguing result using a word analogy prediction formulation and
  hand-selected relations, but the generality of the finding over a
  broader range of lexical relation types and different learning
  settings has not been evaluated. 
In this  paper, we carry out such an evaluation in two learning settings: (1)
  spectral clustering to induce word relations, and (2) supervised
  learning to classify vector differences into relation types. We find
  that word embeddings capture a surprising amount of information,
and that, under suitable
  supervised training, vector subtraction generalises well to a broad
  range of relations, including over unseen lexical items.

\end{abstract}

\section{Introduction}
\label{sec:intro}

Learning to identify lexical relations is a fundamental task in natural
language processing (``\nlp''), and can contribute to many \nlp applications including paraphrasing
and generation, machine translation, and ontology building \cite{Banko+:2007,Hendrickx+:2010}.

Recently, attention has been focused on identifying lexical relations
using 
word embeddings, which are dense, low-dimensional vectors
obtained either from a ``predict-based'' neural network trained to
predict word contexts, or a ``count-based'' traditional distributional
similarity method combined with dimensionality reduction.  The skip-gram
model of \newcite{Mikolov+:2013b} and other 
similar language models have
been shown to perform well on an analogy completion task
\cite{Mikolov+:2013c,Mikolov+:2013,Levy:Goldberg:2014}, in the space of \emph{relational
  similarity} prediction \cite{Turney:2006}, where the task is to
predict the missing word in analogies such as
\lex{A}:\lex{B}\,::\,\lex{C}:\,\lex{\textendash ?\textendash}.
A well-known example involves predicting
the vector \wvec{queen} from the vector combination
$\wvec{king} - \wvec{man} + \wvec{woman}$, where linear operations on
word vectors appear to capture the lexical relation governing the
analogy, in this case 
\relation{opposite-gender}.
  The
results extend to several semantic relations such as
\relation{capital-of}
($\wvec{paris} - \wvec{france} + \wvec{poland} \approx \wvec{warsaw}$)
and morphosyntactic relations such as \relation{pluralisation}
($\wvec{cars} - \wvec{car} + \wvec{apple} \approx \wvec{apples}$).
Remarkably, since the model is not trained for this task, the relational
structure of the vector space appears to be an emergent property.

The key operation in these models is \emph{vector difference}, or
\emph{vector offset}. For example,
the $\wvec{paris} -
\wvec{france}$ vector appears to encode \relation{capital-of},
presumably by cancelling out the features of \wvec{paris} that are
France-specific, and retaining the features that distinguish a capital
city \cite{Levy:Goldberg:2014}.
The success of the simple offset method on analogy completion suggests
that the difference vectors (``\diffvec'' hereafter) must themselves be meaningful: their direction and/or magnitude encodes a lexical relation. 

Previous analogy completion tasks used with 
word embeddings have
limited coverage of lexical relation types. Moreover, the task does not
explore the full implications of \diffvec[s] as meaningful vector space
objects in their own right, because it only looks for a one-best answer
to the particular lexical analogies in the test set. In this paper, we
introduce a new, larger dataset covering many well-known lexical
relation types from the linguistics and cognitive science literature. We
then apply \diffvec[s] to two new tasks: unsupervised and supervised
relation extraction.
First, we cluster the
\diffvec[s] to test whether the clusters map onto true lexical
relations.  
We find that the clustering works remarkably well, although syntactic relations
are captured better than semantic ones.


Second, we perform classification over the \diffvec[s]
and obtain remarkably high accuracy in a
closed-world setting (over a predefined set of word pairs, each of
which corresponds to a lexical relation in the training data). When we
move to an open-world setting including random
word pairs --- many of which do not correspond to any lexical relation in
the training data --- the results are poor. We then investigate
methods for better attuning the learned class representation to the
lexical relations, focusing on methods for automatically synthesising
negative instances. We find that this improves the model performance
substantially.

We also find
that hyper-parameter
optimised count-based methods are competitive with predict-based methods
under both clustering and supervised relation classification, in line
with the findings of \newcite{Levy:2015b}.




\section{Background and Related Work}
\label{sec:background}


A lexical relation is a binary relation $r$ holding between a word
pair $(w_i, w_j)$; for example, the pair \wordpair{cart}{wheel} stands
in the \relation{whole-part} relation.  Relation learning in \nlp
includes relation extraction, relation
classification, and relational similarity prediction.  In relation
extraction, related word pairs in a corpus and the relevant relation
are identified. Given a word pair, the relation classification task
involves assigning a word pair to the correct relation from a
pre-defined set. In the Open Information Extraction paradigm
\cite{Banko+:2007,Weikum:Theobald:2010}, also known as unsupervised relation
extraction, the relations themselves are also learned from the text
(e.g.\ in the form of text labels). On the other hand, relational
similarity prediction involves assessing the degree to which a word
pair \wordpair{A}{B} stands in the same relation as another pair
\wordpair{C}{D}, or to complete an analogy
\lex{A}:\lex{B}\,::\,\lex{C}:\,\lex{\textendash ?\textendash}.  Relation learning is an
important and long-standing task in \nlp and has been the focus of a
number of shared tasks
\cite{Girju:2007b,Hendrickx+:2010,Jurgens+:2012}.

Recently, attention has turned to using vector space models of words for
relation classification and relational similarity prediction. 
Distributional word vectors
have 
been used for detection of relations such as hypernymy
\cite{Geffet:Dagan:2005,Kotlerman+:2010,Lenci:Benotto:2012,Weeds+:2014,Rimell:2014,Santus+:2014}
and qualia structure \cite{Yamada+:2009}.
An exciting development, and the inspiration for this paper, has been
the demonstration that vector difference over 
word embeddings
\cite{Mikolov+:2013} can be used to model word analogy tasks. This has
given rise to a series of papers exploring the \diffvec idea in
different contexts. The original analogy dataset has been used to
evaluate 
predict-based
language models by \newcite{Mnih:Kavukcuoglu:2013} and
also \newcite{Zhila+:2013}, who combine a neural language model with a
pattern-based classifier. 
\newcite{Kim:Marneffe:2013} use word embeddings to derive
representations of adjective scales, e.g.\
\lex{hot---warm---cool---cold}. \newcite{Fu+:2014} similarly use
embeddings to predict hypernym relations, in this case
clustering words by topic to show that hypernym
\diffvec[s] can be broken down into more fine-grained relations. Neural
networks have also been developed for joint learning of lexical and
relational similarity, making use of the WordNet relation hierarchy
\cite{Bordes+:2013,Socher+:2013c,Xu+:2014,Yu:Dredze:2014,Faruqui+:2015,Fried:Duh:2015}.

Another strand of work responding to the vector difference approach has
analysed the structure of 
predict-based embedding models in order to help explain their success on
the analogy and other tasks
\cite{Levy:Goldberg:2014,Levy:Goldberg:2014b,Arora+:2015}. However,
there has been no systematic investigation of the range of relations for
which the vector difference method is most effective, although there
have been some smaller-scale investigations in this direction.
\newcite{Makrai+:2013} divide antonym pairs into semantic classes such
as quality, time, gender, and distance, finding that for about
two-thirds of antonym classes, \diffvec[s] are significantly more
correlated than random.  \newcite{Necsulescu+:2015} train a classifier
on word pairs, using word embeddings to predict coordinates, hypernyms,
and meronyms.  \newcite{Roller:2016} analyse the performance of vector
concatenation and difference on the task of predicting lexical
entailment and show that vector concatenation overwhelmingly learns to
detect Hearst patterns (e.g., \lex{including}, \lex{such as}).
\newcite{Koper+:2015} undertake a systematic study of morphosyntactic
and semantic relations on word embeddings produced with \wordveclong
(``\wordvec'' hereafter; see \secref{sec:embeddings}) for English and
German. They test a variety of relations including word similarity,
antonyms, synonyms, hypernyms, and meronyms, in a novel analogy
task. Although the set of relations tested by \newcite{Koper+:2015} is
somewhat more constrained than the set we use, there is a good deal of
overlap. However, their evaluation is performed in the context of
relational similarity, and they do not perform clustering or
classification on the \diffvec[s].

\section{General Approach and Resources}
\label{sec:datasets}

We define the task of lexical relation learning to
take a set of (ordered) word pairs $\{(w_i, w_j)\}$ and a set of binary
lexical relations $R = \{r_k\}$, and map each word pair $(w_i,
w_j)$ as follows: (a) $(w_i, w_j) \mapsto r_k \in R$, i.e.\
the ``closed-world'' setting, where we assume that all word pairs can be
uniquely classified according to a relation in $R$; or (b)
$(w_i, w_j) \mapsto r_k \in R\, \cup \{\phi\}$ where $\phi$ signifies the fact that
none of the relations in $R$ apply to the word pair in
question, i.e.\ the ``open-world'' setting.

Our starting point for lexical relation learning is the
assumption that important information about various types of relations
is implicitly embedded in the offset vectors. 
While a range of methods have been proposed for composing word vectors
\cite{Baroni+:2012,Weeds+:2014,Roller+:2014}, in this research we focus
exclusively on \diffvec (i.e.\ $\wvec{w_2} - \wvec{w_1}$).
A second assumption is that there exist dimensions, or directions,
in the embedding vector spaces responsible for a particular lexical relation.
Such dimensions could be identified and exploited as part of a clustering 
or classification method, in the context of identifying relations
between word pairs or classes of \diffvec[s]. 

In order to test the generalisability of the \diffvec method,
we require: (1) word embeddings, and (2) a set of lexical relations to evaluate
against. As the focus of this paper is not the word embedding
pre-training approaches so much as the utility of the \diffvec[s]
for lexical relation learning, we take a selection of four pre-trained
word embeddings with strong currency in the literature, as detailed in
\secref{sec:embeddings}. We also include the state-of-the-art
count-based approach of \newcite{Levy:2015b}, to test the
generalisability of \diffvec[s] to count-based word embeddings. 

For the lexical relations, we want a range of
relations that is representative of the types of relational learning tasks 
targeted in the literature, and where there is availability of annotated data. To this end, we construct a
dataset from a variety of sources, focusing on lexical semantic
relations (which are less well represented in the analogy dataset of
\newcite{Mikolov+:2013}), but also including morphosyntactic and morphosemantic relations (see \secref{sec:relations}).

\subsection{Word Embeddings}
\label{sec:embeddings}

\begin{table}
\begin{center}
\begin{tabular}{ccc}
\hline \bf Name & \bf Dimensions  &\bf Training data \\
\hline \wordvec & 300 & $100\times 10^9$\\
\glove  & 200 & $\z\z6\times10^9$\\
\senna & 100 & $\z37\times10^6$\\
\hlbl & 200 & $\z37\times10^6$ \\
\hline
\wordvec[\wiki] & 300 &  $\z50\times10^6$ \\ 
\glove[\wiki] & 300 &  $\z50\times10^6$ \\ 
\svd[\wiki] & 300 &  $\z50\times10^6$ \\
\hline 
\end{tabular}
\caption{The pre-trained word embeddings used in our experiments, with
  the number of dimensions and size of the training
  data (in word tokens). The models
  trained on English Wikipedia (``\wiki'') are in the lower half of the table.}
\label{tab:models-set}
\end{center}
\end{table}

\begin{table*}[!htbp]
\centering
\smaller
\begin{tabular}{lllll} 
\hline \textbf{Relation}  & \textbf{Description} & \textbf{Pairs}& \textbf{Source}& \textbf{Example} \\ \hline
\hyperrel &  hypernym  &    1173  & SemEval'12 + BLESS&\wordpair{\lex{animal}}{\lex{dog}}\\
\merorel &  meronym &  2825  & SemEval'12 + BLESS & \wordpair{\lex{airplane}}{\lex{cockpit}}\\
\attribute &  characteristic quality, action   &  \z\z71  & SemEval'12 & \wordpair{\lex{cloud}}{\lex{rain}} \\
\causepurp &  cause, purpose, or goal  &  \z249  & SemEval'12 & \wordpair{\lex{cook}}{\lex{eat}}\\
\spacerel &  location or time association &  \z235  & SemEval'12 & \wordpair{\lex{aquarium}}{\lex{fish}}\\
\refrel    &  expression or representation &  \z187  & SemEval'12 & \wordpair{\lex{song}}{\lex{emotion}}\\
\eventrel &  object's action &  3583  & BLESS &  \wordpair{\lex{zip}}{\lex{coat}}\\
\nounsp &  plural form of a noun  &  \z100  & MSR & \wordpair{\lex{year}}{\lex{years}} \\
\verbthree &  first to third person verb present-tense form  &  \z\z99  & MSR & \wordpair{\lex{accept}}{\lex{accepts}} \\
\verbpast &  present-tense to past-tense verb form  &  \z100 & MSR & \wordpair{\lex{know}}{\lex{knew}}\\
\verbthreepast &  third person present-tense to past-tense verb form  &  \z100 & MSR & \wordpair{\lex{creates}}{\lex{created}}\\
\lvc &  light verb construction  &  \z\z58  & \newcite{Tan+:2006b}& \wordpair{\lex{give}}{\lex{approval}}\\
\verbnoun &  nominalisation of a verb &  3303  & WordNet & \wordpair{\lex{approve}}{\lex{approval}} \\
\prefix &  prefixing with \lex{re} morpheme  &  \z118  & Wiktionary & \wordpair{\lex{vote}}{\lex{revote}} \\
\nouncoll &  collective noun  &  \z257  & Web source &  \wordpair{\lex{army}}{\lex{ants}}\\
\hline
\end{tabular}
\caption{Description of the 15 lexical relations.}
\label{tab:relation-def}
\end{table*}

We consider four highly successful word embedding models in our experiments:
\wordvec \cite{Mikolov+:2013b,Mikolov+:2013c}, \glove \cite{Pennington+:2014}, \senna
\cite{Collobert:Weston:2008}, and \hlbl
\cite{Mnih:Hinton:2009}, as detailed below.
We also include \svd \cite{Levy:2015b}, a count-based model which
factorises a positive PMI (PPMI) matrix. For consistency of comparison,
we train \svd as well as a version of \wordvec and \glove (which we call \wordvec[\wiki] and \glove[\wiki], respectively) on the
English Wikipedia corpus (comparable in size to the training data of
\senna and \hlbl), and apply the preprocessing of \newcite{Levy:2015b}.
We additionally normalise the \wordvec[\wiki] and \svd[\wiki] vectors to unit length;
\glove[\wiki] is natively normalised by column.\footnote{We ran a series of experiments on
  normalised and unnormalised \wordvec models, and found that
  normalisation tends to boost results over most of our relations (with the exception of
  \eventrel and \nouncoll). We leave a more detailed investigation of
  normalisation to future work.}

\wordvec CBOW (Continuous Bag-Of-Words; \newcite{Mikolov+:2013b}) predicts a word from its context using a model with the objective:
\begin{equation*}
J = \frac{1}{T}\sum_{i=1}^T \log \frac{\exp\left(\wi^\top\sum\limits_{j \in [-c, +c], \\ j\neq 0}{\twij}\right)}{\sum_{k=1}^V {\exp\left(\wk^\top\sum\limits_{j \in [-c, +c], \\ j\neq 0}{\twij}\right)}}
\end{equation*} 
where $\wi$ and $\twi$ are the vector representations for the $i$th
word (as a focus or context word, respectively), $V$ is the vocabulary
size, $T$ is the number of tokens in the corpus, and $c$ is the context
window size.\footnote{In a slight abuse of notation, the subscripts of $\ws$ do double duty, denoting either the embedding for the $i$th token, $\wi$, or $k$th word type,  $\wk$.}
Google News data was used to train the model. We use the focus word vectors, $W = \{ \wk \}_{k=1}^V$, normalised such that each $\|\wk\|=1$.

The \glove model \cite{Pennington+:2014} is based on a similar bilinear formulation, framed as a low-rank decomposition of the matrix of corpus co-occurrence frequencies:
\begin{equation*}
J=\frac{1}{2}\sum_{i,j=1}^V {f(P_{ij})(\wi^\top\twj-\log P_{ij})^2} \, ,
\end{equation*}
 where $w_i$ is a vector for the left context, $w_j$ is a vector for the right context, $P_{ij}$ is the relative frequency of word $j$ in the context of word $i$, and $f$ is a heuristic weighting function to balance the influence of high versus low term frequencies. 
The model was trained on English Wikipedia and the English Gigaword corpus version 5. 

The \svd model \cite{Levy:2015b} uses positive pointwise mutual information (PMI) matrix defined as:
\begin{equation*}
\text{PPMI}(w,c)=\max(\log\frac{\hat{P}(w,c)}{\hat{P}(w)\hat{P}(c)},0) \, ,
\end{equation*}
where $\hat{P}(w,c)$ is the joint probability of word $w$ and context
$c$, and $\hat{P}(w)$ and $\hat{P}(c)$ are their marginal
probabilities. The matrix is factorised by singular value decomposition.

\hlbl \cite{Mnih:Hinton:2009} is a log-bilinear formulation of an $n$-gram language model, which predicts the $i$th word based on context words $(i-n, \ldots, i-2, i-1)$. 
This leads to the following training objective:
\begin{equation*}
J=\frac{1}{T}\sum_{i=1}^T \frac{\exp(\twi^\top\wi+b_i)}{\sum_{k=1}^V{\exp(\twi^\top\wk+b_k)}} \, ,
\end{equation*}
where $\twi=\sum_{j=1}^{n-1}C_j \wij$ is the context embedding, 
$C_j$ is a scaling matrix, and $b_*$ is a bias term. 

The final model, \senna \cite{Collobert:Weston:2008}, was initially
proposed for multi-task training of several language processing tasks,
from language modelling through to semantic role labelling. Here we
focus on the statistical language modelling component, which has a
pairwise ranking objective to maximise the relative score of each word
in its local context:
\begin{align*}
\!\! J=\frac{1}{T} \sum_{i=1}^T \sum_{k =1}^V \max \!\big[ 0, 1&-f(\wsub{i - c}, \ldots, \wsub{i-1}, \wi) \\[-2ex]
 & + f(\wsub{i - c}, \ldots, \wsub{i-1}, \wk) \big] \, ,
\end{align*}  
where the last $c-1$ words are used as context, and $f(x)$ is a non-linear function of the input, defined as a multi-layer perceptron. 

For \hlbl and \senna, we use the pre-trained embeddings from \newcite{Turian+:2010}, trained on
the Reuters English newswire corpus. 
In both cases, the embeddings were scaled by the global standard deviation over the word-embedding matrix, $W_{\text{scaled}}=0.1 \times \frac{W}{\sigma(W)}$. 

For \wordvec[\wiki], \glove[\wiki] and \svd[\wiki] we used English
Wikipedia. We followed the same preprocessing procedure described in
\newcite{Levy:2015b},\footnote{Although the \wordvec model trained
  without preprocessing performed marginally better, we used
  preprocessing throughout for consistency.} i.e., lower-cased all words
and removed non-textual elements. During the training phase, for each
model we set a word frequency threshold of 5. For the \svd model, we
followed the recommendations of \newcite{Levy:2015b} in setting the
context window size to 2, negative sampling parameter to 1, eigenvalue
weighting to 0.5, and context distribution smoothing to 0.75; other
parameters were assigned their default values. For the other models we
used the following parameter values: for \wordvec, context window = 8,
negative samples = 25, hs = 0, sample = 1e-4, and iterations = 15; and
for \glove, context window = 15, x\_max = 10, and iterations = 15.



\subsection{Lexical Relations}
\label{sec:relations}

In order to evaluate the applicability of the \diffvec approach
to relations of different types, we assembled a set of lexical relations in three broad categories: lexical semantic relations, morphosyntactic paradigm relations, and morphosemantic relations.
We constrained the relations to be binary and to have fixed
directionality.{\footnote{Word similarity is not included; it is not easily captured by
    \diffvec since there is no homogeneous ``content'' to the lexical relation which could be captured by the direction and magnitude of a difference vector (other than that it should be small).}
Consequently we excluded symmetric lexical relations such as synonymy.
We additionally constrained the dataset to the words occurring in all
embedding sets.
There is some overlap between our relations and those included in the analogy task of \newcite{Mikolov+:2013}, but we include a much wider range of lexical semantic relations, especially those standardly evaluated in the relation classification literature.
We manually filtered the data to remove duplicates (e.g., as part of
merging the two sources of \hyperrel intances), and normalise directionality.






The final dataset consists of 12,458 triples $\langle\text{relation},
\text{word}_1, \text{word}_2\rangle$, comprising 15 relation types, extracted from
SemEval'12~\cite{Jurgens+:2012}, BLESS~\cite{Baroni:Lenci:2011}, 
the MSR analogy dataset~\cite{Mikolov+:2013}, the light verb dataset of \newcite{Tan+:2006}, Princeton WordNet \cite{Fellbaum:1998}, 
Wiktionary,\footnote{\myurl{http://en.wiktionary.org}}
and a web lexicon of collective nouns,\footnote{\myurl{http://www.rinkworks.com/words/collective.shtml}}
as listed in \tabref{tab:relation-def}.\footnote{The dataset is available at \myurl{http://github.com/ivri/DiffVec}}

\section{Clustering}
\label{sec:clustering}

Assuming \diffvec[s] are capable of capturing all lexical
relations equally, we would expect clustering to be able to identify 
sets of word pairs with high relational similarity, or equivalently clusters of similar offset vectors. Under the
additional assumption that a given word pair corresponds to a unique
lexical relation (in line with our definition of the lexical relation
learning task in \secref{sec:datasets}), a hard clustering approach is
appropriate. In order to test these assumptions, we cluster our 15-relation closed-world dataset in the first
instance, and evaluate against the lexical resources in \secref{sec:relations}.

\begin{figure}
\centering
\input{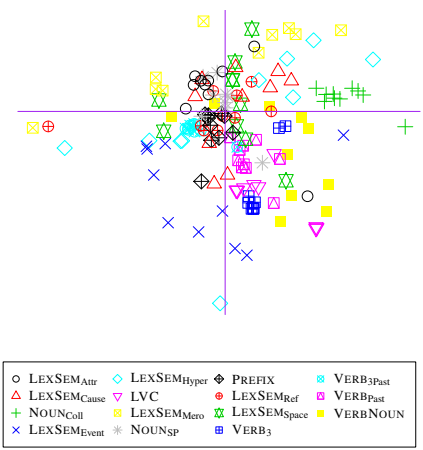}
\caption{t-SNE projection \cite{van2008visualizing} of \diffvec[s] for
    10 sample word pairs of each relation type, based on \wordvec. The intersection of the two axes identify the projection of the zero vector. Best viewed in colour. }
    \label{fig:tsne}
 \end{figure}

As further motivation, we projected the \diffvec space for a small
number of samples of each class using t-SNE \cite{van2008visualizing},
and found that many of the morphosyntactic relations (\verbthree,
\verbpast, \verbthreepast, \nounsp) form tight clusters (\figref{fig:tsne}).



We cluster the \diffvec[s] between all word pairs in our dataset using spectral clustering \cite{Ulrike:2007}. 
Spectral clustering has two hyperparameters: the number of clusters, and the pairwise similarity measure for comparing \diffvec[s]. We tune the hyperparameters over development data, in the form of 15\% of the data obtained by random sampling, selecting the configuration that maximises the V-Measure \cite{Rosenberg:Hirschberg:2007}.
\figref{fig:clustering} presents V-Measure values over the test data for
each of the four word embedding models. We show results for different numbers of clusters, from $N=10$ in
steps of 10, up to $N = 80$ (beyond which the clustering
quality diminishes).\footnote{Although 80 clusters $\gg$ our 15 relation
  types, the SemEval'12 classes each contain
  numerous subclasses, so the larger number may be more realistic.}
Observe that \wordvec achieves the best results, with a V-Measure value
of around 0.36,\footnote{V-Measure returns a value in the range $[0,1]$,
  with 1 indicating perfect homogeneity and completeness.}  which is
relatively constant over varying numbers of clusters.  \glove and \svd mirror
this result, but are consistently below \wordvec at a V-Measure of around
0.31.  \hlbl and \senna performed very similarly, at a substantially
lower V-Measure than \wordvec or \glove, closer to 0.21.
As a crude calibration for these results, over the related clustering task of word sense induction, the best-performing systems in SemEval-2010 Task 4 \cite{Manandahar+:2010} achieved a V-Measure of under 0.2.

\begin{figure}
\centering
\vspace*{-1.2cm}
\input{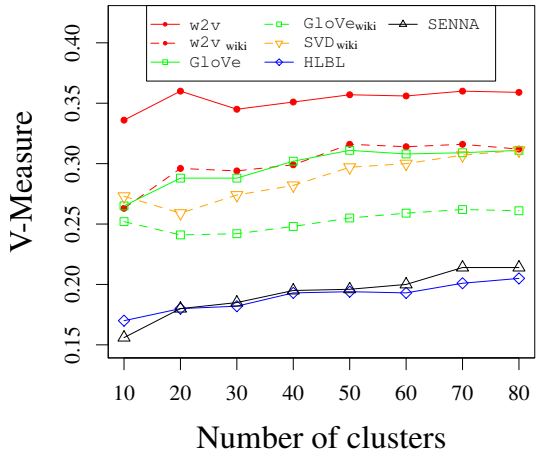}
\caption{Spectral clustering results, comparing cluster quality (V-Measure) and the number of clusters. \diffvec[s] are clustered and compared to the known relation types. Each line shows a different source of word embeddings.}
   \label{fig:clustering}
\end{figure}


The lower V-measure for \wordvec[\wiki] and \glove[\wiki] (as compared to \wordvec and \glove, respectively) indicates that the volume of training data plays a role in the clustering results. However, both methods still perform well above \senna and \hlbl, and \wordvec has a clear
empirical advantage over \glove.
We note that \svd[\wiki] performs almost as well as \wordvec[\wiki], consistent with the results of \newcite{Levy:2015b}. 


\begin{table}[t]
\footnotesize
\centering
\begin{tabular}{lcccc} 
  \hline \bf  & \bf \wordvec & \bf \glove & \bf \hlbl & \bf \senna \\
  \hline
\attribute     &  \bst 0.49  &       0.54 &       0.62 &       0.63 \\
\causepurp     &  \bst 0.47  &       0.53 &       0.56 &       0.57 \\
\spacerel      &  \bst 0.49  &       0.55 &       0.54 &       0.58 \\
\refrel        &  \bst 0.44  &       0.50 &       0.54 &       0.56 \\
\hyperrel      &       0.44  &       0.50 & \bst  0.43 &       0.45 \\
\eventrel      &  \bst 0.46  &       0.47 &       0.47 &       0.48 \\
\merorel       &  \bst 0.40  &       0.42 &       0.42 &       0.43 \\
\nounsp        &  \bst 0.07  &       0.14 &       0.22 &       0.29 \\
\verbthree     &  \bst 0.05  &       0.06 &       0.49 &       0.44 \\
\verbpast      &  \bst 0.09  &       0.14 &       0.38 &       0.35 \\
\verbthreepast &       0.07  &  \bst 0.05 &       0.49 &       0.52 \\
\lvc           &  \bst 0.28  &       0.55 &       0.32 &       0.30 \\
\verbnoun      &  \bst 0.31  &       0.33 &       0.35 &       0.36 \\
\prefix        &       0.32  &  \bst 0.30 &       0.55 &       0.58 \\
\nouncoll      &  \bst 0.21  &       0.27 &       0.46 &       0.44 \\
\hline
\end{tabular}
\caption{The entropy for each lexical relation over the clustering
  output for each set of pre-trained word embeddings.}
\label{tab:entropy}
\end{table}

We additionally calculated the entropy for each lexical
relation, based on the distribution of instances
belonging to a given relation across the different
clusters (and simple MLE). For each embedding
method, we present the entropy for the cluster size
where V-measure was maximised over the development
data. Since the samples are distributed nonuniformly,
we normalise entropy results for each
method by $\log(n)$ where $n$ is the number of samples
in a particular relation. The results are in \tabref{tab:entropy},
with the lowest entropy (purest clustering) for each
relation indicated in bold.

Looking across the different lexical relation types,
the morphosyntactic paradigm relations (\nounsp
and the three \verbrel relations) are by far the easiest to capture.
The lexical semantic relations,
on the other hand, are the hardest to capture
for all embeddings.

Considering \wordvec embeddings, for \verbthree there was a single cluster consisting
of around 90\% of \verbthree word pairs. 
Most errors resulted from POS ambiguity, leading to confusion
with \verbnoun in particular. Example \verbthree
pairs incorrectly clustered 
are: \wordpair{study}{studies}, \wordpair{run}{runs}, 
and \wordpair{like}{likes}.
This polysemy results in the distance represented
in the \diffvec for such pairs being above
average for \verbthree, 
and 
consequently clustered with other cross-POS relations.

For \verbpast, a single relatively pure cluster
was generated, with minor contamination due to
pairs
such as \wordpair{hurt}{saw},
\wordpair{utensil}{saw}, and \wordpair{wipe}{saw}. Here, the noun \lex{saw}
is ambiguous with a high-frequency past-tense verb;
{\it hurt} and {\it wipe} also have ambigous POS.

A related phenomenon was observed for
\nouncoll , where the instances were assigned to a
large mixed cluster containing word pairs where
the second word referred to an animal, reflecting the fact that
most of the collective nouns in our dataset relate
to animals, e.g.\ \wordpair{stand}{horse}, \wordpair{ambush}{tigers},
\wordpair{antibiotics}{bacteria}. This is interesting from a
\diffvec point of view, since it shows that the
lexical semantics of one word in the pair can
overwhelm the semantic content of the \diffvec (something that we return
to investigate in \secref{sec:lexmem}). \merorel
was also split into multiple clusters
along topical lines, with separate clusters
for weapons, dwellings, vehicles, etc.




Given the encouraging results from our clustering experiment, we next
evaluate \diffvec[s] in a supervised relation classification setting.


\section{Classification}
\label{sec:classification1}

A natural question is whether we can accurately characterise lexical
relations through supervised learning over the
\diffvec[s]. 
For these experiments we use the \wordvec, \wordvec[\wiki], and \svd[\wiki]
embeddings exclusively (based on their superior performance in the
clustering experiment), 
and a subset of the relations which is both representative of the
breadth of the full relation set, and for which we have sufficient data for supervised training and evaluation, namely:
\nouncoll, \eventrel, \hyperrel, \merorel, \nounsp, \prefix, \verbthree, \verbthreepast, and \verbpast (see \tabref{tab:relation-def}).

We consider two applications: (1) a \closedworld setting similar to the
unsupervised evaluation, in which the classifier only encounters 
word pairs which correspond to one of the nine relations; and (2) a more challenging \openworld setting where random
word pairs --- which may or may not correspond to one of our
relations --- are included in the evaluation.
For both settings, we further investigate whether there is a lexical memorisation effect for a broad range of relation types of the sort identified by \newcite{Weeds+:2014} and \newcite{Levy:2015} for hypernyms, by experimenting with disjoint training and test vocabulary.

\subsection{\closedworld Classification}
\label{sec:classclosed}

For the \closedworld setting, we train and test a multiclass classifier
on datasets comprising 
$\langle \diffvec, r \rangle$ pairs, where $r$ is one of our nine
relation types, and \diffvec is based on one of \wordvec,
\wordvec[\wiki] and \svd. As a baseline, we cluster the data as described in
\secref{sec:clustering}, running the clusterer several times over the
9-relation data to select the optimal V-Measure value based on the
development data, resulting in 50 clusters. We label each cluster with
the majority class based on the training instances, and evaluate the
resultant labelling for the test instances.

We use an SVM with a linear kernel, and report results from 10-fold cross-validation in \tabref{mclass-table}.

The SVM achieves a higher F-score than the baseline on almost every
relation, particularly on \hyperrel, and the lower-frequency \nounsp,
\nouncoll, and \prefix.  Most of the relations --- even the most
difficult ones from our clustering experiment --- are classified with
very high F-score.
That is, with a
simple linear transformation of the embedding dimensions, we are able to
achieve near-perfect results.
The \prefix relation achieved markedly lower recall, resulting in a lower F-score, 
due to large
differences in the predominant usages associated with the respective words
(e.g., \wordpair{union}{reunion}, where the vector for \lex{union} is
heavily biased by contexts associated with trade unions, but
\lex{reunion} is heavily biased by contexts relating to social
get-togethers; and \wordpair{entry}{reentry}, where \lex{entry} is
associated with competitions and entrance to schools, while
\lex{reentry} is associated with space travel).
Somewhat surprisingly, given the small dimensionality of the input
(vectors of size 300 for all three methods), we found that the linear SVM slightly
outperformed a non-linear SVM using an RBF kernel. 
We observe no real difference between \wordvec[\wiki] and \svd[\wiki],
supporting the hypothesis of \newcite{Levy:2015b} that under appropriate
parameter settings, count-based methods achieve high results. The impact
of the training data volume for pre-training of the embeddings is also
less pronounced than in the case of our clustering experiment.

\begin{table}[t]
\centering
\footnotesize
\begin{tabular}{lcccc}
  \hline
  \textbf{Relation} & \textbf{Baseline} & \textbf{\wordvec} & \textbf{\wordvec[\wiki]} & \textbf{\svd[\wiki]}\\
   \hline
\hyperrel     & 0.60 & 0.93 & 0.91 & 0.91  \\ 
\merorel      & 0.90 & 0.97 & 0.96 & 0.96  \\ 
\eventrel     & 0.87 & 0.98 & 0.97 & 0.97  \\ 
\nounsp       & 0.00 & 0.83 & 0.78 & 0.74  \\ 
\verbthree    & 0.99 & 0.98 & 0.96 & 0.97  \\ 
\verbpast     & 0.78 & 0.98 & 0.98 & 0.95  \\
\verbthreepast& 0.99 & 0.98 & 0.98 & 0.96  \\ 
\prefix       & 0.00 & 0.82 & 0.34 & 0.60  \\ 
\nouncoll     & 0.19 & 0.95 & 0.91 & 0.92  \\ 
\hline
Micro-average &  0.84 & 0.97 & 0.95  & 0.95 \\
\hline
\end{tabular}
\caption{F-scores (\fscore) for
  \closedworld classification, for a baseline method based on clustering
  + majority-class labelling, a multiclass linear SVM trained on \wordvec, \wordvec[\wiki] and \svd[\wiki] \diffvec inputs.}
\label{mclass-table}
\end{table}

\subsection{\openworld Classification}
\label{sec:open-world}

We now turn to a more challenging evaluation setting: a test set
including word pairs drawn at random. 
This setting aims to illustrate whether a \diffvec-based classifier is capable of 
differentiating related word pairs from noise, and can be applied to
open data to learn new related word pairs.\footnote{Hereafter we provide results for \wordvec only, as we found that \svd achieved similar results.}

For these experiments, we train a binary classifier for each relation type, using $\frac{2}{3}$ of our relation data for training and $\frac{1}{3}$ for testing. 
The test data is augmented with an equal quantity of random pairs, generated as follows: 
\begin{compactenum}[(1)]
\item sample a seed lexicon by drawing words proportional to
  their frequency in Wikipedia;\footnote{Filtered to 
consist 
of words for which we have embeddings.} 
\item take the Cartesian product over pairs of words from the seed lexicon; 
\item sample word pairs uniformly from this set.
\end{compactenum}
This procedure generates word pairs that are representative of the
frequency profile of our corpus.

We train 9 binary RBF-kernel SVM classifiers on the training
partition, and evaluate on our randomly augmented test set.
Fully annotating our random word pairs is prohibitively expensive, so
instead, we manually annotated only the word pairs which were
positively classified by one of our models.
The results of our experiments are presented in the left half of
\tabref{mclassrand-table}, in which we report on results over the combination of the
original test data from \secref{sec:classclosed} and the random word pairs, noting that
recall (\recall) for \openworld takes the form of relative recall
\cite{Pantel+:2004} over
the positively-classified word pairs.
The results are much lower than for the closed-word setting
(\tabref{mclass-table}), most notably in terms of precision (\precision). 
For instance, the random pairs \wordpair{have}{works}, \wordpair{turn}{took}, and \wordpair{works}{started}
were incorrectly classified as \verbthree, \verbpast and \verbthreepast,
respectively. 
That is, the model captures syntax, but lacks the ability to capture
lexical paradigms, and tends to overgenerate. 

\subsection{\openworld Training with Negative Sampling}

\label{sec:open-world-noise}

\begin{table}[t]
\centering
\footnotesize
\begin{tabular}{lc@{\,\,\,\,}c@{\,\,\,\,}c@{\,\,\,\,}cc@{\,\,\,\,}c@{\,\,\,\,}c}
  \hline
  \multirow{2}{*}{\textbf{Relation}} & \multicolumn{3}{c}{\textbf{Orig}}
  && \multicolumn{3}{c}{\textbf{$+$neg}}\\
  \cline{2-4}
  \cline{6-8}
  & \precision & \recall & \fscore && \precision & \recall & \fscore\\ 
  \hline
\hyperrel      & 0.95   &  \bst 0.92 & \bst0.93 && \bst 0.99  &  0.84 & 0.91      \\
\merorel       & 0.13   &  \bst 0.96 & 0.24 && \bst 0.95  &  0.84 & \bst 0.89 \\ 
\eventrel      & 0.44  &  \bst 0.98 & 0.61 &&  \bst 0.93  &   0.90 & \bst 0.91 \\ 
\nounsp        & 0.95   &  \bst 0.68 & 0.8 && \bst 1.00  &  \bst 0.68  & \bst 0.81 \\ 
\verbthree     & 0.75   &   \bst 1.00  & 0.86 && \bst 0.93  &  0.93 & \bst 0.93 \\ 
\verbpast      & 0.94   &  \bst 0.86 & \bst 0.90 && \bst 0.97  &  0.84 & \bst 0.90   \\
\verbthreepast & 0.76   &  \bst 0.95 & 0.84&& \bst 0.87  & 0.93  & \bst 0.90 \\ 
\prefix        & \bst 1.00   &  \bst 0.29 & \bst 0.44 && \bst 1.00  &  0.13 & 0.23 \\ 
\nouncoll      & 0.43   &  \bst 0.74 & 0.55 && \bst 0.97  &  0.41 & \bst 0.57    \\ 
\hline
\end{tabular}
\caption{\label{mclassrand-table} Precision (\precision) and recall (\recall) for
  \openworld classification, using the binary classifier without (``Orig'')
  and with (``$+$neg'') negative samples .}
\end{table}

To address the problem of incorrectly classifying random word pairs as
valid relations, we retrain the classifier on a dataset comprising both
valid and automatically-generated negative distractor samples.
The basic intuition behind this approach is to construct samples which
will force the model to learn decision boundaries that more tightly
capture the true scope of a given relation.
To this end, we automatically generated two types of negative distractors:
\begin{compactdesc} 
\item[opposite pairs:] generated by switching the order of word pairs,
  \oppos~=~$\wvec{word_1}-\wvec{word_2}$. This ensures the classifier
  adequately captures the asymmetry in the relations.
\item[shuffled pairs:] generated by replacing $w_2$ with a random word $w^\prime_2$
  from the same relation,
  \shuff~=~$\wvec{word^\prime_2}-\wvec{word_1}$. 
  This is targeted at relations that take specific word classes
  in particular positions, e.g., \wordpair{\postag{VB}}{\postag{VBD}} word
  pairs, so that the model learns to encode the relation rather than simply learning the properties of the
  word classes.
\end{compactdesc} 
Both types of distractors are added to the training set, such that there are equal numbers of valid relations, opposite pairs and shuffled pairs.





After training our classifier, we evaluate its predictions in the same
way as in \secref{sec:open-world}, using the same test set combining
related and random word pairs.\footnote{But noting that relative
  recall for the random word pairs is based on the pool of positive
  predictions from both models.}
The results are shown in the right half of \tabref{mclassrand-table} (as ``$+$neg'').  
Observe that the precision is much higher and recall somewhat lower compared to the classifier trained with only positive samples.
This follows from the adversarial training scenario: using negative
distractors results in a more conservative classifier, that correctly
classifies the vast majority of the random word pairs as not corresponding to
a given relation, resulting in higher precision at the expense of a
small drop in recall.
Overall this leads to  higher F-scores, as shown in
\figref{fig:f-pos-neg}, other than for hypernyms (\hyperrel) and prefixes (\prefix).
For example, the standard classifier for \nouncoll learned to match word pairs including an animal
name (e.g., \wordpair{\lex{plague}}{\lex{rats}}), while training with
negative samples resulted in much more conservative predictions and
consequently much lower recall. 
The classifier was able to capture \wordpair{herd}{horses}
but not \wordpair{\lex{run}}{\lex{salmon}},
\wordpair{\lex{party}}{\lex{jays}} or
\wordpair{\lex{singular}}{\lex{boar}} as instances of \nouncoll,
possibly because of polysemy.
The most striking difference in performance was for \merorel, where the
standard classifier generated many false positive noun pairs (e.g.\
\wordpair{\lex{series}}{\lex{radio}}), but the false positive rate was
considerably reduced with negative sampling. 

\begin{figure}[t]
\centering
  \includegraphics[width=0.9\columnwidth]{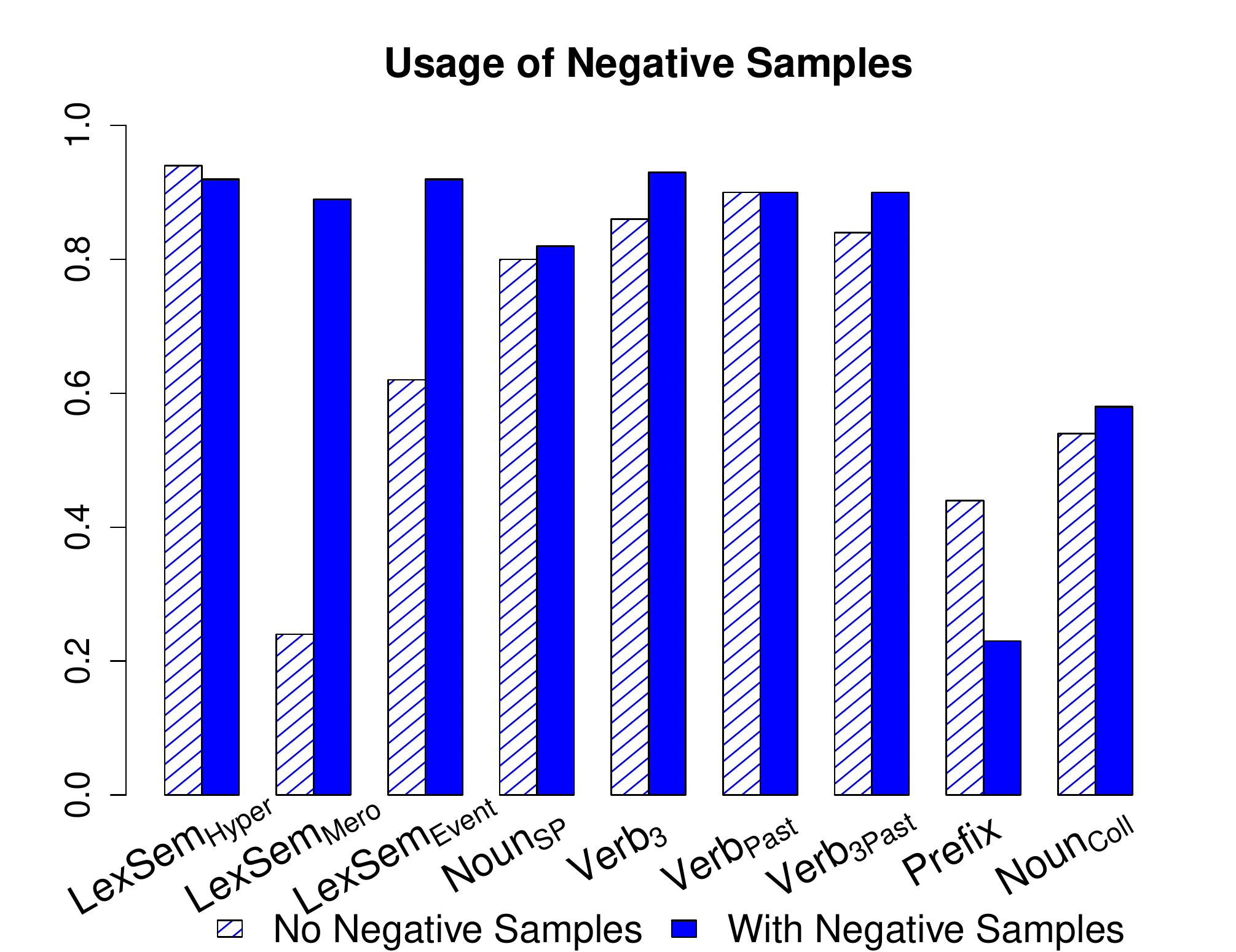}
  \caption{F-score for \openworld classification, comparing models
    trained with and without negative samples.}
  \label{fig:f-pos-neg}
\end{figure}

\subsection{Lexical Memorisation}
\label{sec:lexmem}

\newcite{Weeds+:2014} and \newcite{Levy:2015} recently showed that
supervised methods using {\diffvec}s achieve artificially high results
as a result of ``lexical memorisation'' over frequent words associated
with the hypernym relation. For example, \wordpair{animal}{cat},
\wordpair{animal}{dog}, and \wordpair{animal}{pig} all share the
superclass \lex{animal}, and the model thus learns to classify as positive any word
pair with \lex{animal} as the first word. 

To address this effect, we follow \newcite{Levy:2015} in splitting our vocabulary into
training and test partitions, to ensure there is no overlap between
training and test vocabulary. We then train
classifiers with and without negative sampling
(\secref{sec:open-world-noise}), incrementally adding the random word
pairs from \secref{sec:open-world} to the test data (from no random word
pairs to five times the original size of the test data) to investigate the interaction of
negative sampling with greater diversity in the test set when there is a
split vocabulary. The results are shown in \figref{fig:lex-mem}. 

\begin{figure}
\centering
\input{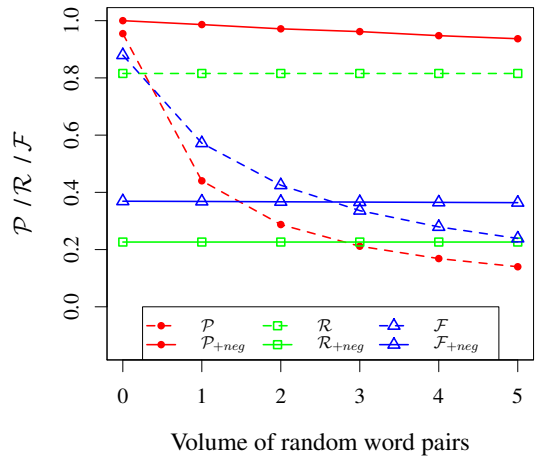}
\caption{Evaluation of the \openworld model when trained on split
    vocabulary, for varying numbers of random word pairs in the test
    dataset (expressed as a multiplier relative to the number of
    \closedworld test instances).}
   \label{fig:lex-mem}
\end{figure}


Observe that the precision for the standard classifier decreases rapidly as more random word pairs are added to the test data. In comparison, the precision when negative sampling is used shows only a small drop-off, indicating that negative sampling is effective at maintaining precision in an \openworld setting even when the training and test vocabulary are disjoint. This benefit comes at the expense of recall, which is much lower when negative sampling is used (note that recall stays relatively constant as random word pairs are added, as the vast majority of them do not correspond to any relation). At the maximum level of random word pairs in the test data, the F-score for the negative sampling classifier is higher than for the standard classifier.
 
\section{Conclusions}
\label{sec:conclusions}

This paper is the first to test the generalisability of the vector
difference approach across a broad range of lexical relations (in raw
number and also variety). 
Using clustering we showed that many types of
morphosyntactic and morphosemantic differences are captured by \diffvec[s], but that lexical semantic relations are captured less well,
a finding which is consistent with previous work \cite{Koper+:2015}.
In contrast, classification over the \diffvec[s] works
extremely well in a closed-world setting, showing that dimensions of
\diffvec[s] encode lexical relations.
Classification performs less well over open data, although with the introduction of automatically-generated negative
samples, the results improve substantially. Negative sampling also improves classification when the training and test vocabulary are split to minimise lexical memorisation.
Overall, we conclude that the \diffvec approach has
impressive utility over a broad range of lexical relations, especially
under supervised classification.


\section*{Acknowledgments}

LR was supported by EPSRC grant EP/I037512/1 and ERC Starting Grant
DisCoTex (306920). TC and TB were supported by the Australian Research Council.

	
\bibliographystyle{acl2016}
\bibliography{strings,collections,papers}

\end{document}